\documentclass{article}

\usepackage[preprint]{neurips_2023}

\usepackage[utf8]{inputenc} % allow utf-8 input
\usepackage[T1]{fontenc}    % use 8-bit T1 fonts
\usepackage{hyperref}       % hyperlinks
\usepackage{url}            % simple URL typesetting
\usepackage{booktabs}       % professional-quality tables
\usepackage{amsfonts}       % blackboard math symbols
\usepackage{nicefrac}       % compact symbols for 1/2, etc.
\usepackage{microtype}      % microtypography
\usepackage{xcolor}         % colors
\usepackage{multirow}

\usepackage{amsmath}
\usepackage{amssymb}
\usepackage{mathtools}
\usepackage{amsthm}
\usepackage{natbib}
\setcitestyle{numbers,square}
\title{Anti-aliasing Predictive Coding Network \\ for Future Video Frame Prediction}

\author{%
  Chaofan Ling \\
  South China University of Technology\\
  \texttt{wichaofan@mail.scut.edu.cn} 
   \And
   Weihua Li \\
   South China University of Technology \\
   \texttt{whlee@scut.edu.cn} \\
   \AND
   Junpei Zhong \\
   The Hong Kong Polytechnic University \\
   \texttt{joni.zhong@polyu.edu.hk} \\

}

\begin{document}

\maketitle

\begin{abstract}
  We introduce here a predictive coding based model that aims to generate accurate and sharp future frames. Inspired by the predictive coding hypothesis and related works, the total model is updated through a combination of bottom-up and top-down information flows, which can enhance the interaction between different network levels. Most importantly, We propose and improve several artifacts to ensure that the neural networks generate clear and natural frames. Different inputs are no longer simply concatenated or added, they are calculated in a modulated manner to avoid being roughly fused. The downsampling and upsampling modules have been redesigned to ensure that the network can more easily construct images from Fourier features of low-frequency inputs. Additionally,  the training strategies are also explored and improved to generate believable results and alleviate inconsistency between the input predicted frames and ground truth. Our proposals achieve results that better balance pixel accuracy and visualization effect.
\end{abstract}

\section{Introduction}
\label{introduction}

Future video frame prediction has been rapidly developed recently \cite{wu2021greedy, chang2021mau, chang2022strpm, gao2022simvp, wu2022optimizing, ye2023video}. This looking-ahead ability could be applied to various fields including autonomous driving \cite{morris2008learning, wei2010prediction}, robotic systems \cite{finn2017deep, Gao2021Future}, rainfall forecasting \cite{shi2017deep, shi2015convolutional} etc. In particular, this self-supervised learning method for video representation can also be migrated to downstream tasks \cite{han2019video, wang2020self}, to solve the difficulties of label acquisition in supervised learning tasks.

Despite its enormous value in terms of application, predicting an accurate and clear video frame remains a challenging task due to the uncertainty and curse of dimensionality involved. Existing works \cite{villegas2017decomposing, oliu2018folded, wang2018eidetic, lin2020motion, chang2021mau} still suffer from lack of appearance details, low prediction accuracy, and high computational overhead. We observe that most of the works usually employ a similar unidirectional end-to-end architecture \cite{villegas2017decomposing, oliu2018folded, wang2018eidetic, chang2021mau, chang2022strpm, gao2022simvp, ye2023video}. For instance, VPTR \cite{ye2023video} first encodes historical frames into high-level representations, then uses transformer network for prediction, and finally decodes and reconstructs the pixels. It only predicts high-level representations in semantic space, potentially leading to a loss of prediction details, which can manifest as predicted images being blurred or inconsistent with actual motion trajectories finally. The convolutional LSTMs (ConvLSTM) solve the problems well \cite{shi2015convolutional, wang2018eidetic}, it directly calculates video frames and performs predictions at each level. However, the memory state is still only updated within each Convolutional LSTM level, relying less on the hierarchical visual features of the other network levels.

The predictive coding framework \cite{rao1999predictive, han2017rhythms, aitchison2017or, teufel2020forms} provides a novel and effective way to solve the above problems. It is updated through a combination of bottom-up and top-down information flows to enhance the interaction between different network levels, and uses prediction errors to achieve effective feedback connections. One of the most typical predictive coding model is the PredNet proposed by Lotter et al. \cite{lotter2017deep}, which strictly follows the computational manner of traditional predictive coding framework and achieves state-of-the-art results at that time. 
%Similarly, Straka et al. uses ConvLSTM to construct PreCNet \cite{straka2023precnet}, in which they seem to only change the direction of the information flow compared with PredNet, that is, the predictions are propagated upward while the prediction errors are propagated downward. The authors did not report the reasons. 
However, the PPNet \cite{ling2022pyramidal} proposes different ideas according to recent cognitive science and neuroscience hypotheses, e.g., the information being propagated up the hierarchy isn't just limited to prediction errors, but also includes signals such as sensory input \cite{teufel2020forms}. The model's unique temporal pyramid architecture enables longer-term predictions at higher levels with lower update frequencies, broadening the temporal receptive field of higher-level neurons and reducing computational overhead.  Our model is inspired by this idea, but differs in several ways.

We first observed the inconsistencies in its temporal signal processing in PPNet. Specifically, the sensory input propagated to higher level is lagged, which exacerbates as the network level increases. Furthermore, the sensory input and prediction error are simply concatenated for calculation, which forces the predictive unit at higher-level to predict the lower level prediction error simultaneously. It is both contradictory and challenging. We will expose and analyze these issues in detail and redesign the computational specification in Section 2.

Additionally, how to construct clear and natural video frames is not well considered by most works including PPNet \cite{ling2022pyramidal, su2020convolutional, gao2022simvp, tan2022temporal, straka2023precnet}. For instance, the Conv-TT-LSTM \cite{su2020convolutional} reports pretty good pixel-wise accuracy (usually evaluated using Structural Similarity SSIM \cite{wang2004image} and Peak Signal-to-Noise Ratio PSNR \cite{hore2010image}), while its visual performance seems unsatisfactory. Besides, the SimVP \cite{gao2022simvp} and TAU \cite{tan2022temporal} achieve higher SSIM and PSNR scores, however, they do not report the evaluation on perceptual metric LPIPS \cite{zhang2018unreasonable} nor show adequate visualization results. We identify two sources that contribute to the problem: 1) using a deterministic loss such as mean squared error (MSE) forces the model to prioritize reducing the overall error, which usually resulting in generating blurred images (especially in complex and unpredictable scenes); 2) current downsampling or upsampling artifacts are not aggressive in suppressing aliasing. Several works \cite{lee2018stochastic, babaeizadeh2018stochastic, jin2020exploring, ling2022predictive} propose to use adversarial training to sharpen the generated images. However, adversarial training may cause the generated frames to deviate from the actual trajectory, which needs to be carefully controlled.

We find that directly using deep features to compute the loss without adversarial training is also effective for image sharpening.
We expose that the insufficient representation ability of deterministic loss or metric cause the first problem described above, which only measure differences at pixel-level.
The LPIPS \cite{zhang2018unreasonable} also points out that the scores of MSE, SSIM and PSNR may not align with the human visual system. Therefore, it proposes to use neural networks with stronger representation capabilities to measure the perceptual similarity between images. In this work, we use the pre-trained model provided by LPIPS to compute the perceptual loss, and the results demonstrate the superiority of this approach. Compared with adversarial training, its training is more stable and energy efficient, while avoiding the predicted images from deviating from the actual trajectory.

To further address the aliasing cause by non-ideal downsampling and upsampling filters, we propose to employ low-pass filters with learnable cutoff frequencies. According to the Shannon-Nyquist signal processing framework \cite{shannon1949communication}, in order to restore the signal without distortion, the sampling frequency should be greater than twice the highest frequency in the signal spectrum. Typically, a low-pass filter is used to remove high-frequency signals above half the sampling frequency, which can be ignored.  However, selecting an appropriate cutoff frequency can be challenging. The proportion of high-frequency signals that need to be suppressed may vary depending on the scenarios, input size or even the feature maps. To address this issue, we propose define the cutoff frequencies as learnable parameters, to enable the model adaptively choose which signals need to be attenuated. Additionally, the upsampling artifact has also been redesigned. It does not modify the continuous representation, its sole purpose is to increase the output sampling rate. Once aliasing is adequately suppressed, the model will be forced to implement more natural hierarchical refinement, thus generating more realistic and natural-looking images.

Finally, input inconsistency between training and testing in temporal sequential tasks remains a challenging topic. The predicted frames usually serve as new inputs to enable continuous prediction. In this work, we propose to calculate the ``Encoding Loss" during training to alleviate the impact of prediction error accumulation during long-term predictions, which measures the Euclidean distance between the higher-level representation of the predicted frame and the ground truth. This forces network modules to extract features from ``imperfect" predicted frames that are as similar as possible to the ground truth. Ideally, the features of the predicted frame encoded by the encoding module are exactly the same as those of the ground truth, then the input inconsistency is finally resolved. Our project is available at \url{https://github.com/ANNMS-Projects/PNVP}

\begin{figure*}[!t]
	
	\centering{\includegraphics[width=1.0\textwidth]{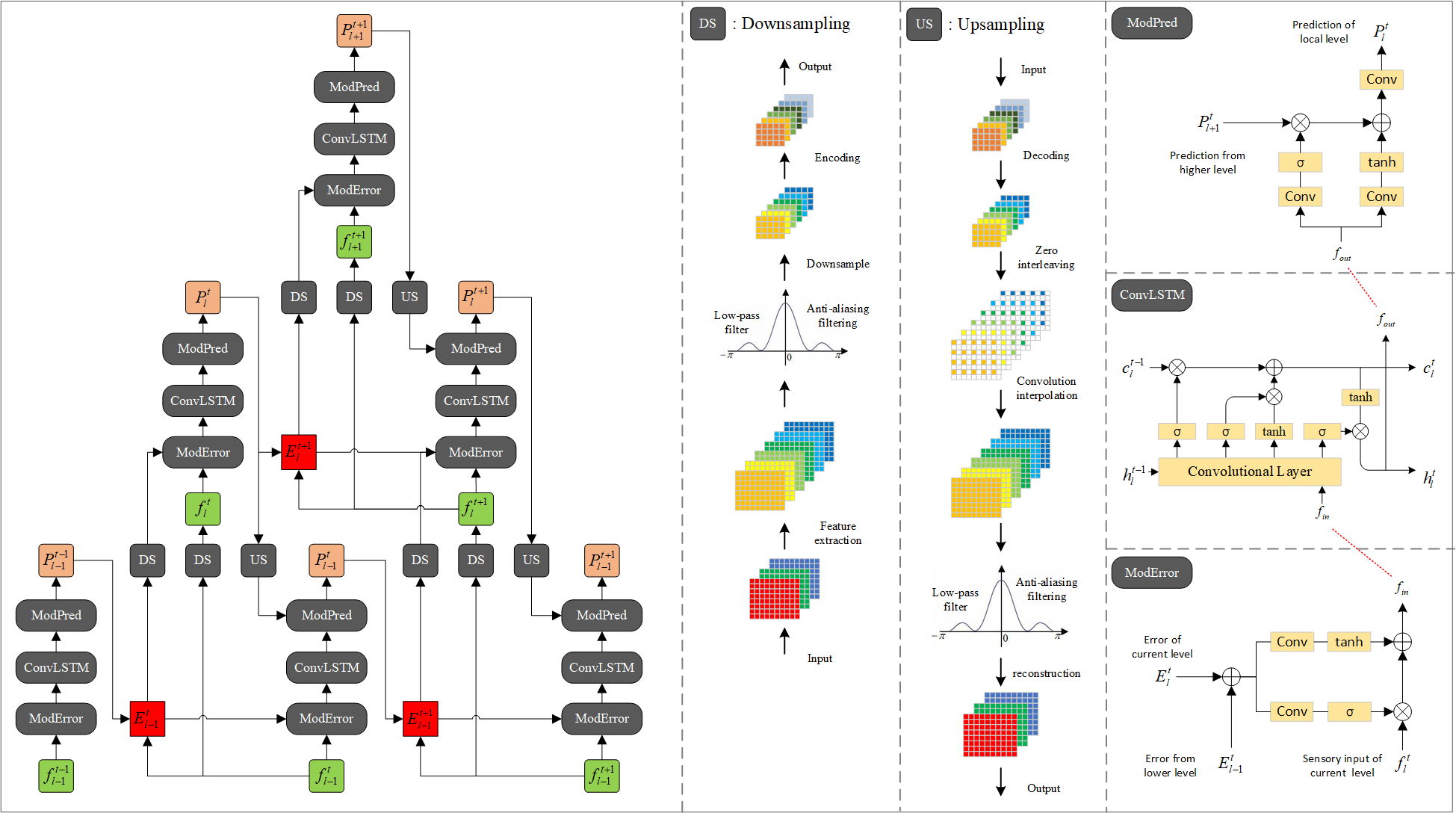}}
	%\vskip -0.05in
	\caption{Overview of our model. From left to right are the overall framework, the downsampling artifact, the upsampling artifact and the detailed calculation process. "ModError" and "ModPred" are special modulation modules. $f_l^t$, $P_l^t$ and $E_l^t$ represent the sensory input, prediction and prediction error at time-step $t$ and level $l$, respectively. Zoom in for a better view.}
	
	\label{fig:TotalNet}
	\vskip -0.1in
\end{figure*}

\section{The overall architecture}

Figure \ref{fig:TotalNet} shows the overall framework of our model. In the video frame prediction task, $f_l^t$ denotes the representation of frame $x_t$ at level $l$. The predictive unit,  composed of ``ModError", ``ConvLSTM" and ``ModPred" modules, combines $f_l^t$ with other signals such as prediction from higher level and prediction error from lower level, to calculate the local prediction $P_l^t$. The $P_l^t$ is then propagated downward to lower level for calculating prediction $P_{l-1}^t$ of lower-level representation. Moreover, it is also utilized to compare with the sensory input $f_l^{t+1}$ of next time step to calculate the prediction error $E_l^{t+1}$, which reflects the current prediction performance and is used to correct the calculation of the network at the next time step. The design of the overall architecture is inspired by the temporal pyramid framework of PPNet \cite{ling2022pyramidal}, but it differs in several ways.

\begin{figure*}[!t]
	\centering{\includegraphics[width=1.0\textwidth]{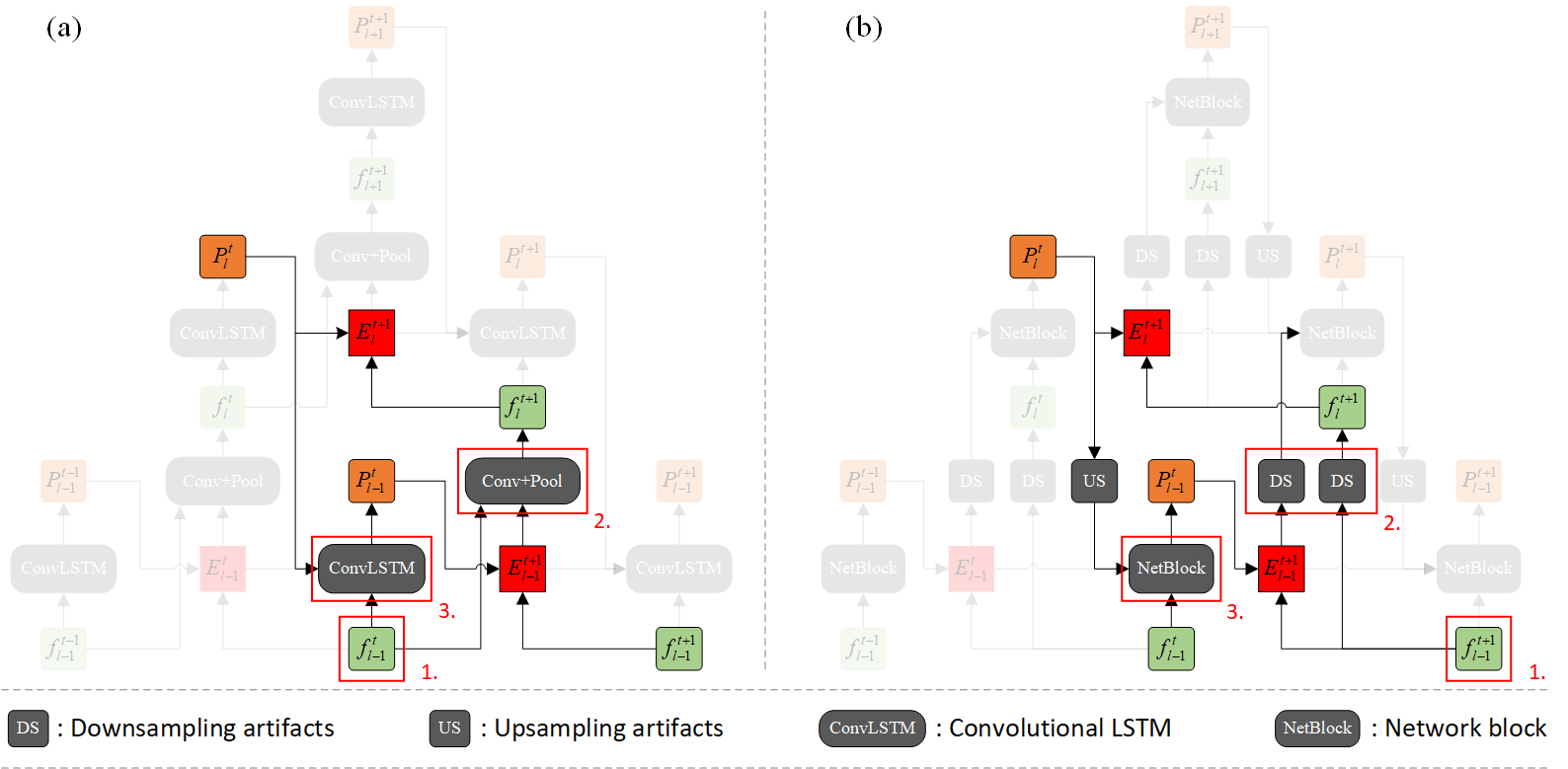}}
	\vskip -0.05in
	\caption{Differences between our model and PPNet.(a): The PPNet. (b): our model. We make several changes which are shown in the red boxes. The ``NetBlock" indicates the predictive unit of our model, which is composed of ``ModError", ``ConvLSTM" and ``ModPred" shown in Fig \ref{fig:TotalNet}.}
	\label{fig:PPNV1_PPNV2}
	\vskip -0.1in
\end{figure*}

Firstly, we find that the sensory input propagated to higher level is lagged in PPNet, which exacerbates as the network level increases. As illustrated in Figure \ref{fig:PPNV1_PPNV2} (a), the sensory input $f_l^{t+1}$ is calculated from the combination of $f_{l-1}^t$ and prediction error $E_{l-1}^{t+1}$, while $P_l^t$ is considered to be a prediction of $f_l^{t+1}$. In other words, the $P_l^t$ is considered to predict $f_{l-1}^{t}$ and $E_{l-1}^{t+1}$. However, the $P_l^t$ is eventually propagated downward and combined with $f_{l-1}^t$ to make the lower-level prediction, so what $P_t^l$ predicts is lag and meaningless. Particularly, this delay exacerbates as the network level increases. For instance, the sensory information contained in $f_{l+1}^{t+1}$ (the third level in Fig \ref{fig:PPNV1_PPNV2} (a)) is actually derived from $f_{l-1}^{t-1}$. Therefore, to address this problem, we propose propagating up the sensory input $f_{l-1}^{t+1}$ (Fig \ref{fig:PPNV1_PPNV2} (b) 1.) of next time step  instead of $f_{l-1}^t$, which matches in the time dimension.

Secondly, in PPNet, the higher level sensory input $f_l^{t+1}$ represents both lower-level signals $f_{l-1}^t$ and $E_{l-1}^{t+1}$, which means that the $P_l^t$ needs to predict the relevant information of prediction error $E_{l-1}^{t+1}$ at the same time. Paradoxically, the prediction error $E_{l-1}^{t+1}$ is calculated from $P_{l-1}^t$ and $f_{l-1}^{t+1}$, while the generation of $P_{l-1}^t$ is inseparable from the higher-level prediction $P_l^t$. In short, $P_l^t$ predicts the $E_{l-1}^{t+1}$, while the $E_{l-1}^{t+1}$ must be generated by $P_l^t$ first, which is contradictory. To address the problem, we propose to calculate the prediction error and sensory input separately. As shown in Fig \ref{fig:PPNV1_PPNV2} (b), the higher-level sensory input $f_l^{t+1}$ only represents $f_{l-1}^{t+1}$ at lower-level, while the prediction error is excluded. 

\section{Improve and redesign the artifacts}

\subsection{The modulation module}

Multiple different signals are simply concatenated and fed into the ConvLSTM unit for subsequent calculation in PPNet. Unfortunately, repeated multiplication of huge weight matrices is likely to cause the gradients to vanish or explode. It further causes substantial increase in the amount of parameters and calculation overhead since the ConvLSTM unit needs to calculate four outputs. 
To address the problem, we propose to introduce a modulation module before and after the ConvLSTM  unit for signal preprocessing and postprocessing (the ``ModError" and ``ModPred"). This novel module is designed to better fuse different signals and alleviate the difficulty with gradient propagation. 

Traditional approaches for fusing two different inputs usually involve dimension-wise concatenation or point-wise addition. However, the former may cause the gradients to vanish or explode due to their huge weight matrices, while the latter may roughly destroy the original distribution if the signals differ greatly.
The modulation module calculates in a modulated manner, which can prevent different signals from being directly fused. 
This idea is also inspired by the attention mechanism: 
%the brain responds strongly when predictions are severely inconsistent with the environment \cite{friston2003learning, friston2008hierarchical, hohwy2008predictive}.
%For ``ModError" module, the  operation can be considered as the process of attention attachment and transfer.
% According to Clark et al. \cite{aitchison2017or}, 
the larger the prediction error, the more attention is given \cite{friston2003learning, friston2008hierarchical, hohwy2008predictive, aitchison2017or}. Therefore, we propose that the prediction error can be viewed as an attention matrix, and the attachment and transfer of attention is equivalent to the process of scaling and shifting the sensory input.

\begin{figure*}[!t]
	\centering{\includegraphics[width=1.0\textwidth]{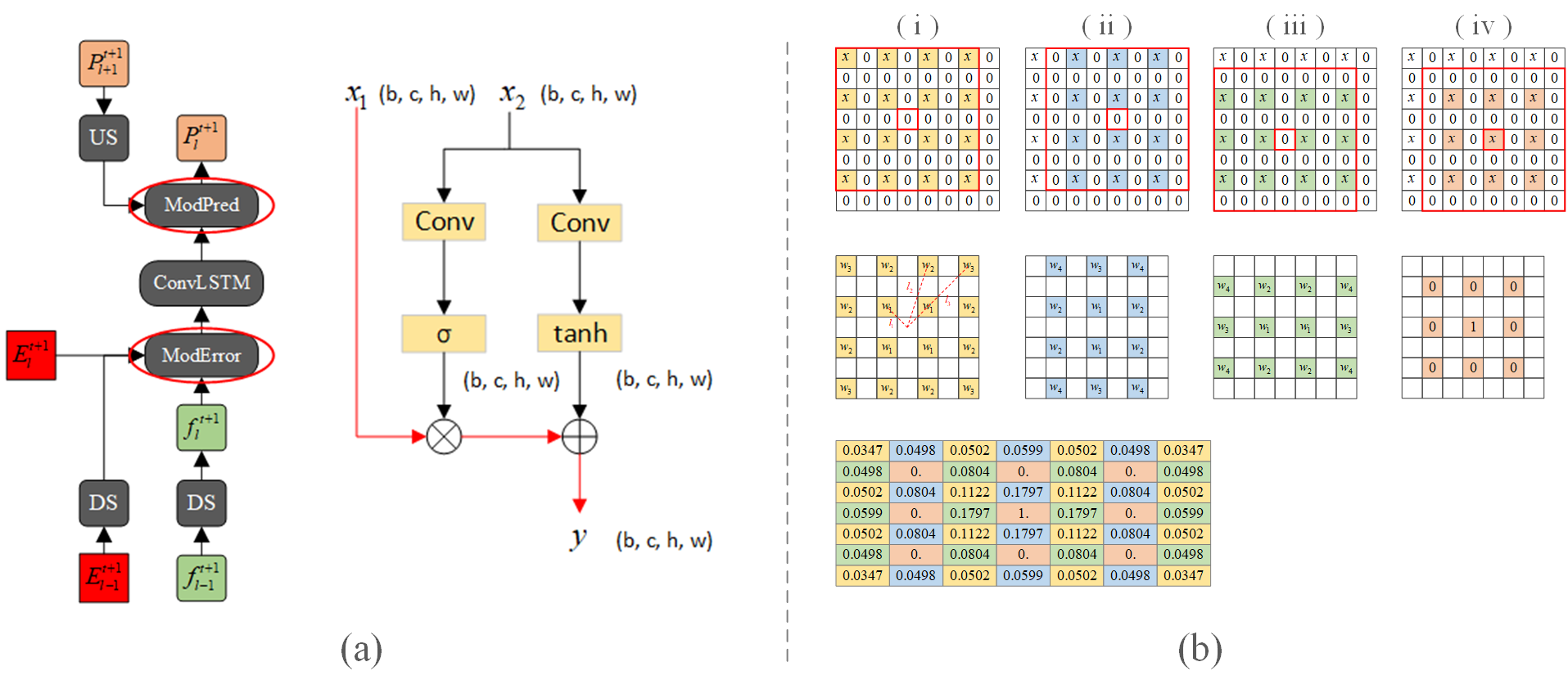}}
	\vskip -0.05in
	\caption{(a): The structure and calculation of modulation module. (b): The interpolation calculation during upsampling, where the first row shows the features involved in 4 different cases in a $2 \times 2$ box
		\begin{math}
			\left (
			\begin{smallmatrix}
				0&0\\0&x
			\end{smallmatrix}
			\right )
	\end{math}, the second row shows the convolutional kernel parameters involved and their initialization calculation, and the third row shows the complete initialized kernel. Zoom in for a better view.}
	\label{fig:MergeStyle}
	\vskip -0.1in
\end{figure*}

The specific calculation is shown in Eq.\ref{eq:modulate} and Figure\ref{fig:MergeStyle} (a), it distinguishes the primary signal $x_1$ and the auxiliary signal $x_2$ first. In this work, the sensory input containing more information that effectively describes the current scene is selected as the main signal, since the neural network preferentially learns low-frequency features \cite{xu2019training, xu2019frequency}. Next, the auxiliary signal  $x_2$ is convolved separately to obtain the scaling and shifting matrices ($m_{sc}$ and $m_{sf}$), 	where $f(\cdot)$ represents a typical convolutional unit and $\theta$ denotes the parameter. The role of $sigmoid$ and $tanh$ is to constrain the matrices between (0, 1) and (-1, 1), respectively. The final output $y$ is obtained by scaling and shifting the primary signal $x_1$ with matrices  $m_{sc}$ and $m_{sf}$. The $\alpha$ is a learnable coefficient that enables the scaling matrix limited between (0, 1) by the $sigmoid$ to obtain an amplification function. Since the calculation of signal $x_1$ only involves point-wise multiplication and addition, so the gradient is more stable.
%\vspace{-0.05in}
\begin{equation}
	m_{sc} = sigmoid \ f(x_2, \theta_{sc}), \quad
	m_{sf} = tanh \ f(x_2, \theta_{sf}), \quad
	y = \alpha \cdot m_{sc} \cdot x_1 + m_{sf}
	\label{eq:modulate}
\end{equation}
%\vspace{-0.15in}

Similarly, in the post-processing stage, we also use the modulation module to process the signals. In this work, we take the higher-level prediction as the primary signal $x_1$ and the output of the ConvLSTM as the auxiliary signal $x_2$. This is because the decoded features from higher levels tend to be more helpful for the final output. Furthermore, since gradient backpropagation of decoded signals can be more challenging, treating them as primary signals $x_1$ can alleviate this problem.

\subsection{The downsampling artifact}

Traditional approaches for downsampling, such as pooling and regular convolution, are not aggressive in suppressing aliasing. Aliasing, a subtle and critical issue, has recently attracted attention \cite{zou2022delving, vasconcelos2021impact, karras2021alias}. It usually manifests as high-frequency signals mixing with low-frequency signals during sampling,  resulting in incoherent sampled signals. According to the Shannon-Nyquist signal processing framework \cite{shannon1949communication}, to address the problem, a high-quality low-pass filter is required. 

In this work, we use the Hamming Window \cite{madisetti1997digital} to design the low-pass filter, which is described as Eq.\ref{eq:hamming}, where $N$ denotes the length of filter. The calculation of coefficients is shown in Eq.\ref{eq:cf}, where $f_c$ and $f_s$ represent the cut-off and sampling frequency respectively. 
%As what we mentioned in Section \ref{introduction}, the cutoff frequency $f_c$ should be between 0 and half the sampling frequency $f_s / 2$, , and it will be limited to (0, 1) during calculation, which is achieved by $2f_c / f_s$.
 Finally we get the low-pass filter by multiplying the Hamming Window and the coefficients: $h(n) = \lambda(n) \cdot w(n)$. 
\vspace{-0.3cm}

\begin{equation}
	w(n) = 0.54 - 0.46 \ cos(\frac{2 \pi n}{N} ), \quad 0 \le n \le N-1
	\label{eq:hamming}
\end{equation}

\begin{equation}
	\lambda (n) = f_c \frac{2}{f_s} \ sinc (f_c \frac{2}{f_s} (n - 0.5(N-1))) 
	\label{eq:cf}
\end{equation}

\begin{table*}[!b]
    \vskip -0.2in
	\centering
	\renewcommand\arraystretch{1.4}
	\caption{The average and standard deviation of the learnable ratio $2f_c / f_s$ at each network level.}
	
	\vskip 0.1in
	
	\resizebox{\linewidth}{!}
	{
	\begin{tabular}{lcccccccccc}
		\toprule
		\multirow{2}{*}{Levels} & \multicolumn{5}{c}{Average}                & \multicolumn{5}{c}{Standard Deviation}      \\
		& \uppercase\expandafter{\romannumeral1}      & \uppercase\expandafter{\romannumeral2}       & \uppercase\expandafter{\romannumeral3}       & \uppercase\expandafter{\romannumeral4}       & \uppercase\expandafter{\romannumeral5}       & \uppercase\expandafter{\romannumeral1}       & \uppercase\expandafter{\romannumeral2}       & \uppercase\expandafter{\romannumeral3}       & \uppercase\expandafter{\romannumeral4}       & \uppercase\expandafter{\romannumeral5}        \\ \cmidrule(r){1-11}
		Caltech                          & 0.8619 & 0.8803 & 0.8748 & 0.5602 & 0.5000 & 0.0212 & 0.0042 & 0.0143 & 0.0852 & 2.78e-8 \\
		KITTI                            & 0.8346 & 0.8738 & 0.8863 & 0.7964 & 0.5086 & 0.0353 & 0.0063 & 0.0067 & 0.0833 & 0.0392 \\ \bottomrule
	\end{tabular}
}
\end{table*}
Specifically, we use low-pass filter of length $N=25$ and resize it to a 5 $\times$ 5 convolution kernel. In general, the larger the filter size, the higher the filtering quality, but it also requires more computational overhead. Most importantly, we propose to define the cut-off frequencies as learnable parameters $f_c^i, \ 1 \le i \le C$ for each input feature map, where $C$ denotes the number of channels. It allows the model to select appropriate attenuation degree through training. Exactly, what we define is the ratio $2f_c / f_s$ instead of the specific cut-off frequency value. According to what we mentioned in Section \ref{introduction}, the cutoff frequency $f_c$ should be between 0 and half the sampling frequency $f_s / 2$,  so the ratio coefficient $2f_c / f_s$ will be constrained to (0, 1) with $sigmoid$ function. 
Finally, we calculate the parameters for the low-pass filters according to Eq. \ref{eq:hamming} and Eq. \ref{eq:cf}. 
In this way, we can obtain a convolution kernel of shape $(C, 1, 5, 5)$, which is depth-separable that only performs anti-aliasing filtering and downsampling (stride = 2). 

It turns out that the above proposal is necessary according to the results shown in Table 1. We counted the average and standard deviation of the ratio coefficients $2f_c / f_s$ learned at different network levels, where Table 1 shows the results obtained by training on the Caltech \cite{Dollar2012PAMI} and KITTI \cite{Geiger2013IJRR} datasets. It can be observed from the table that the ratio coefficients exhibit significant variation with network levels, datasets and input feature maps. Therefore, it is difficult to manually determine an appropriate ratio coefficient for effective anti-aliasing.

\begin{figure*}[!t]
	\centering{\includegraphics[width=1.0\textwidth]{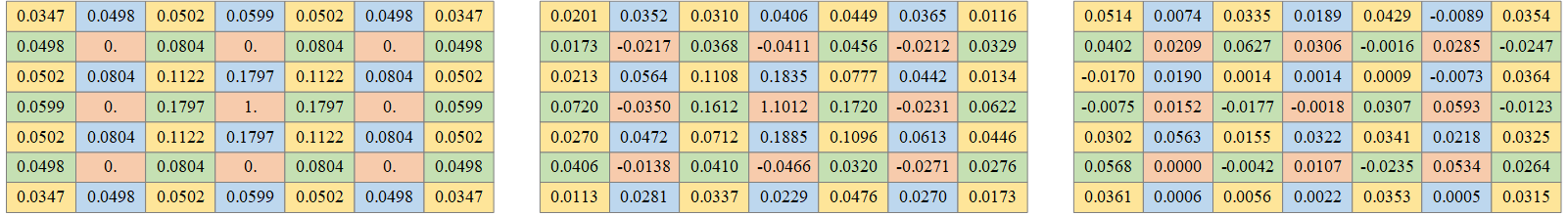}}
	\vskip -0.05in
	\caption{Left: the initialized matrix; Middle: the final learned matrix using the proposed initialization method; Right: the final learned matrix using the Kaiming uniform initialization.}
	\label{fig:initialize}
	\vskip -0.1in
\end{figure*}

\subsection{The upsampling artifact}

Inspired by styleGAN3 \cite{karras2021alias}, ideal upsampling should not modify the continuous representation, its sole purpose is to increase the output sampling rate. Similarly, our upsampling operation consists of two steps: we first increase the sampling rate to $s_{out} = s_{in} \cdot m$ by interleaving $m-1$ zeros between each input sample ($m$ is usually set to 2). The feature map after staggered filling with zeros is shown in Fig \ref{fig:MergeStyle} (b), where $x$ refers to the original input features. Next, we perform convolution interpolation on the output and then low-pass filtering (Fig \ref{fig:TotalNet} ``US").

In this work, we replace the bilinear or bicubic $2\times$ upsampling filter with a $7\times7$ depthwise separable convolution approximation, which is initialized in a special way to achieve the interpolation calculation function. Specificlly, we use the weighted average approach to initialize the parameters, in which the weight of the feature $x$ is inversely proportional to its distance. In $2\times$ upsampling, we have to consider the four computational cases shown in Fig \ref{fig:MergeStyle} (b), where we focus on the calculation at different positions in the $2 \times 2$ box
\begin{math}
	\left (
	\begin{smallmatrix}
		0&0\\0&x
	\end{smallmatrix}
	\right )
\end{math}
. There are different convolution parameters and input features $x$  involved in the calculation of each case. 

Taking the first calculation (Fig \ref{fig:MergeStyle} (b) (\romannumeral1)) as an example, we need to calculate the interpolation of 0 in the center. There are three kinds of features with different distances around the center point, and we will assign different weights to these features. Firstly, the weight is inversely proportional to the distance, which is shown as Eq.\ref{eq:proportion}, where $w_1, w_2, w_3$ represent the weights under the three kinds of distances $l_1, l_2, l_3$ respectively. Secondly, the sum of all weights is set to 1 (Eq.\ref{eq:sum}), so that the weights can be calculated (Eq.\ref{eq:w}). The parameters for cases of (\romannumeral2) and (\romannumeral3) are calculated in the same way. For the last case (\romannumeral4), we expect to preserve the original feature $x$, so we set the central parameter to 1 and the others to 0. Finally, the complete initialized kernel is shown in the third row of Fig \ref{fig:MergeStyle} (b).
\begin{equation}
	w_1 : w_2 : w_3 = \frac{1}{l_1}:\frac{1}{l_2}:\frac{1}{l_3} = \frac{1}{\sqrt{2}}:\frac{1}{\sqrt{3^2+1} }:\frac{1}{\sqrt{3^2+3^2}}
	\label{eq:proportion}
\end{equation}
\begin{equation}
	4 w_1 + 8 w_2 + 4 w_3 = 1
	\label{eq:sum}
\end{equation}
\begin{equation}
	w_1=0.1122 \ ; \quad w_2 = 0.0502 \ ; \quad w_3=0.0374
	\label{eq:w}
\end{equation}
The proposed initialization can better guide the learning of the network. Figure \ref{fig:initialize} shows the final matrices learned by using different initialization methods. They are obtained by averaging the 64 convolution kernels at level \uppercase\expandafter{\romannumeral1}. It can be observed that the matrix obtained with the proposed initialization is generally consistent with the initial data, which means that it is indeed performing the interpolation calculation. The parameters are only slightly adjusted to accommodate the distribution of input data. The matrix obtained by Kaiming Uniform \cite{he2015delving} initialization is completely different. We suspect that it is performing an unknown decoding process rather than interpolation calculation. Our ablation studies (Figure \ref{fig:ablation}) also shows that better results are obtained using the proposed initialization.

\section{The training strategy}

The training strategy plays an important role in generating high-quality predicted frames. Specifically, we calculate three kinds of losses including prediction loss $\mathcal{L}_1$, encoding loss $\mathcal{L}_2$ and LPIPS loss $\mathcal{L}_{lpips}$ during training. Firstly, the prediction loss $\mathcal{L}_1$ is indispensable which measures the squared Euclidean distance between local prediction $P_l^{t-1}$ and next sensory input $f_l^t$. Its calculation is expressed as Eq.\ref{losses1}, where $L$ denotes the number of network level, $\lambda_t$ and $\lambda_l$ represent the weighting coefficient at time-step $t$ and level $l$ respectively. $T = T_1 + T_2$ denotes the total length of input sequence, where $T_1$ is the length of ground truth sequence and $T_2$ is the length of predicted sequence.
\begin{equation}
	\mathcal{L}_1 = \sum_{t=0}^{T} \sum_{l=0}^{L} \lambda_t \cdot \lambda_l \cdot (f_l^t - P_l^{t-1})^2, \quad
	\mathcal{L}_2 = \sum_{t=T_1}^{T} \sum_{l=1}^{L} \lambda_t \cdot \lambda_l \cdot (f_l^t - \hat{f}_l^t)^2, 
	\label{losses1}
\end{equation}

The encoding loss $\mathcal{L}_2$ is calculated to alleviate the inconsistency between ground truth frames and predicted frames. 
Long-term prediction is a important task for future video frame prediction. Predicted frames instead of ground truth frames are used to enable continuous prediction, however,
the predicted frames are ``imperfect'', which will cause the predictions to becoming increasingly blurry as prediction errors accumulate. Therefore, we propose to calculate the encoding loss $\mathcal{L}_2$ to force the network modules to extract features from ``imperfect" predicted frames that are as similar as possible to the ground truth frames. The specific calculation is shown in Eq.\ref{losses1}, where $f_l^t$ and $\hat{f}_l^t$ indicate the representation of the ground truth frame and predicted frame at time-step $t$ and level $l$, respectively. They are obtained by calculating with the same encoding module.
\begin{equation}
	\mathcal{L}_{lpips} = \sum_{t=0}^{T} \lambda_t  \cdot f_{lpips}(f^t, P^{t-1}), \quad \quad
	\mathcal{L}_{total} = \mathcal{L}_{1} + \mathcal{L}_{2} + \lambda \cdot \mathcal{L}_{lpips}
	\label{loss:total}
\end{equation}
The LPIPS loss $\mathcal{L}_{lpips}$ plays an important role in image sharpening, which is described as Eq.\ref{loss:total}. The $f_{lpips}$ denotes the LPIPS pretrained model (VGG as backbone), $P^{t-1}$ denotes the predicted frame generated at time-step $t-1$ and $f_t$ is the target frame. Deterministic losses such as mean square error or Euclidean distance usually calculate differences between images at pixel-wise level, which may cause the model to generating specious and ambiguous results (according to Figure \ref{fig:ablation}, the model generates increasingly blurred images when LPIPS loss is not utilized, however, it still maintains high pixel-wise level accuracy (SSIM and PSNR)). Therefore, we suggest to simultaneously employ deep features to measure the differences between images, which is achieved by a neural network, to generate more believable results. Finally, the total loss $\mathcal{L}_{total}$ is defined as the sum of the above losses (Eq.\ref{loss:total}). The $\lambda$ is a coefficient to increase the proportion of LPIPS loss.

\section{Results}

\subsection{Evaluation with state-of-the-art}
Similar to previous works, we use SSIM, PSNR and LPIPS for quantitative evaluation. The LPIPS score is calculated using the designated pre-trained model (Alex network as the backbone). Higher scores of SSIM and PSNR and lower score of LPIPS indicate better results. We validate the proposed methods on several popular datasets, which are preprocessed in the same way as previous works \cite{ling2022pyramidal, straka2023precnet, lin2020motion, lotter2017deep} to ensure the fairness.

\begin{figure*}[!t]
	\centering
	\begin{minipage}{0.45\linewidth}
		\centering
		\vspace{-0.35cm}
		\includegraphics[width=\linewidth]{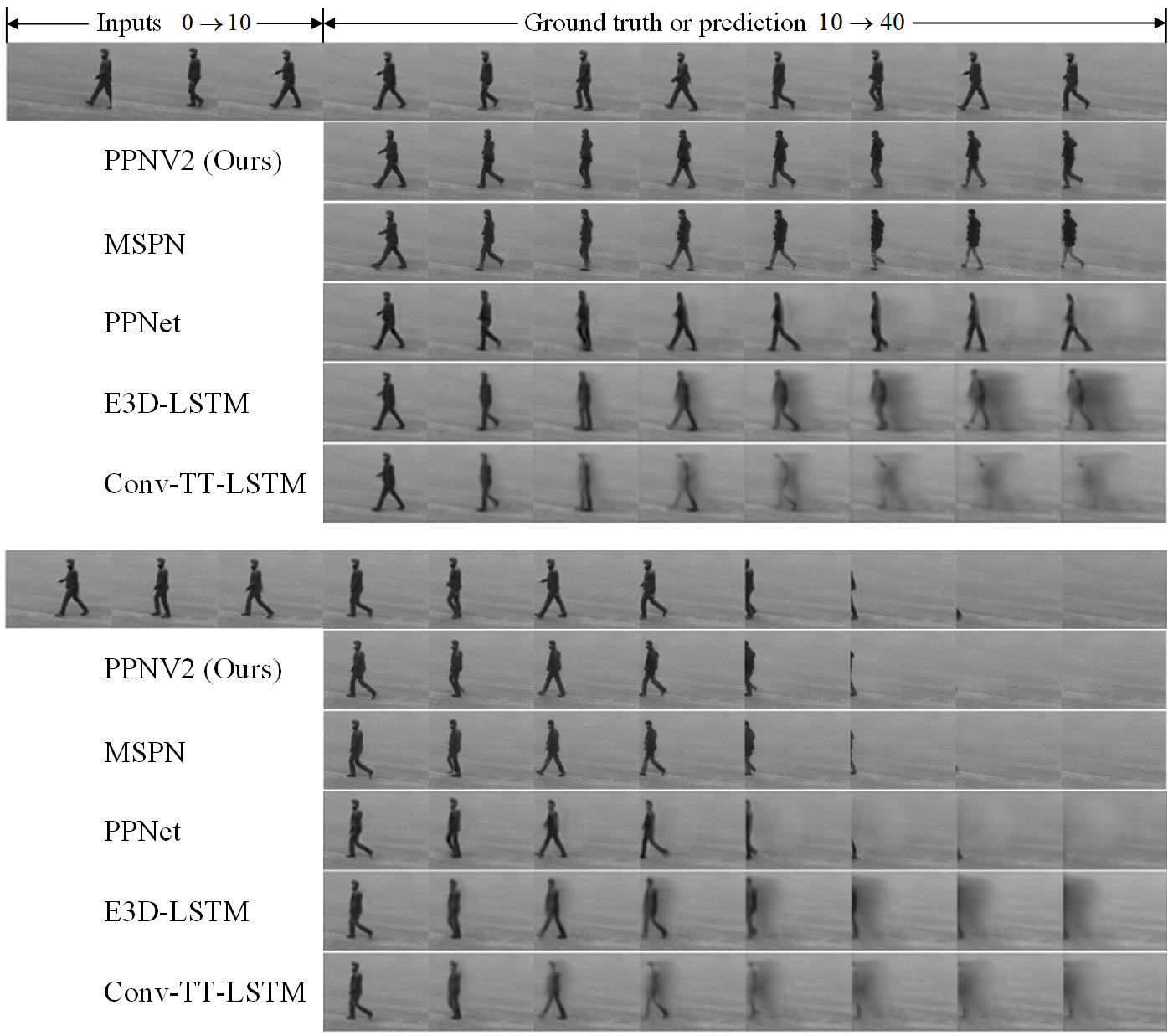}
	\end{minipage}
	\quad
	%\hfill
	\begin{minipage}{0.51\linewidth}
		\centering
		\vspace{-0.28cm}
		\renewcommand\arraystretch{1.44}
		\small
		\resizebox{\linewidth}{!}
		{
			\begin{tabular}{lcccccc}
				\hline
				\multirow{2}{*}{Methods} & \multicolumn{3}{c}{10 $\rightarrow$ 20} & \multicolumn{3}{c}{10 $\rightarrow$ 40} \\
				& SSIM $\uparrow$  & PSNR $\uparrow$  & LPIPS $\downarrow$ 
				& SSIM $\uparrow$  & PSNR $\uparrow$  & LPIPS $\downarrow$  \\
				\hline
				MCNet  \cite{villegas2017decomposing}                  & 0.804  & 25.95  & -     & 0.73   & 23.89  & -      \\
				fRNN   \cite{oliu2018folded}                  & 0.771  & 26.12  & -     & 0.678  & 23.77  & -      \\
				PredRNN  \cite{wang2017predrnn}                & 0.839  & 27.55  & -     & 0.703  & 24.16  & -      \\
				PredRNN++   \cite{wang2018predrnn++}             & 0.865  & 28.47  & -     & 0.741  & 25.21  & -      \\
				VarNet  \cite{jin2018varnet}                 & 0.843  & 28.48  & -     & 0.739  & 25.37  & -      \\
				SAVP-VAE \cite{lee2018stochastic}  & 0.852  & 27.77 & 8.36 &0.811  & 26.18  & \textcolor{blue}{11.33} \\
				E3D-LSTM \cite{wang2018eidetic}                & 0.879  & 29.31  & -     & 0.810  & 27.24  & -      \\
				STMF  \cite{jin2020exploring}                 & 0.893  & 29.85  & 11.81 & 0.851  & 27.56   & 14.13  \\
				Conv-TT-LSTM  \cite{su2020convolutional}             & \textcolor{blue}{0.907}  & 28.36  & 13.34 & 0.882  & 26.11  & 19.12 \\
				LMC-Memory  \cite{lee2021video}                 & 0.894  & 28.61 & 13.33 & 0.879  & 27.50   & 15.98  \\
				PPNet  \cite{ling2022pyramidal}                 & 0.886  & 31.02  & 13.12 & 0.821  & 28.37  & 23.19  \\
				MSPN \cite{ling2022predictive}      &  0.881      &  31.87     & 7.98     &  0.831      &    28.86    &   14.04    \\ 
				VPTR \cite{ye2023video}      &  0.859      &  26.13     & \textcolor{blue}{7.96}     &  -     &    -    &   -    \\ 
				SimVP \cite{gao2022simvp}      &  0.905      &  \textcolor{blue}{33.72 }     & -    &  \textcolor{blue}{0.886}      &    \textcolor{blue}{32.93}    &   -    \\ 
				TAU \cite{tan2022temporal}      &  \textcolor{red}{0.911}      &  \textcolor{red}{34.13 }     & -    &  \textcolor{red}{0.897}      &    \textcolor{red}{33.01}    &   -    \\ \hline
				Ours       &  0.893      &  32.05     & \textcolor{red}{4.76}     &  0.833      &    28.97    &   \textcolor{red}{8.93}  \\
				\hline
			\end{tabular}
		}
	\end{minipage}
	\caption{Evaluation results on the KTH dataset. The left shows the visualization examples of 30 predicted frames. The right shows quantitative results, where the metrics are averaged over the 10 or 30 predicted frames. Red and Blue indicate the best and second best results, respectively.}
	\vskip -0.05in
	\label{fig:KTH}
\end{figure*}

Figure \ref{fig:KTH} provides the visualization examples and quantitative results on the KTH dataset. Our approach balances the pixel-wise accuracy and visualization effect well, that is, it generates natural and clear frames while maintaining high SSIM and PSNR scores. In contrast, the Conv-TT-LSTM \cite{su2020convolutional} reports fairly high pixel-wise accuracy, while its visual performance seems to be unsatisfactory. This confirms our previous conjecture that using only deterministic losses like mean squared error may lead to specious results. The SimVP \cite{gao2022simvp} and TAU \cite{tan2022temporal} achieve even more surprisingly high pixel-wise accuracy, however, they do not report the evaluation on perceptual metric LPIPS nor show the visualization examples. On the contrary, some works such as SAVP-VAE  \cite{lee2018stochastic} and VPTR \cite{ye2023video}  obtain  better performance on LPIPS, while their pixel-wise accuracy is reduced.
\begin{figure}[!t]
	\centering{\includegraphics[width=1.0\textwidth]{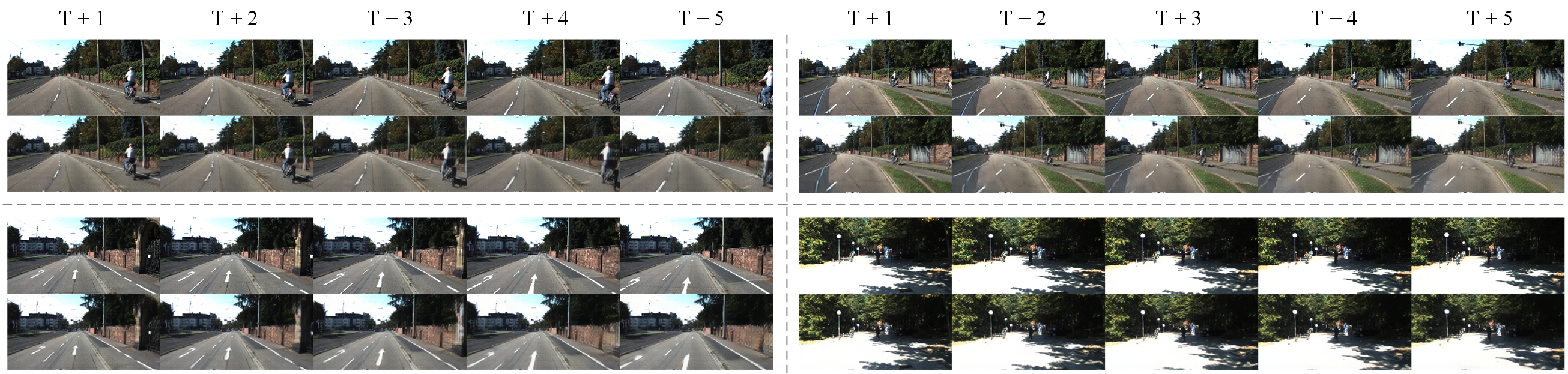}}
	\vspace{-0.5cm}
	\caption{Visualization examples on the KITTI dataset. In each group, the first row indicate the ground truth while the second row is prediction. Zoom in for a better view.}
	\label{fig:KITTI}
	%\vskip -0.1in
\end{figure}
\begin{figure*}[!t]
	\centering
	\begin{minipage}{0.45\linewidth}
		\centering
		\vspace{-0.30cm}
		\includegraphics[width=\linewidth]{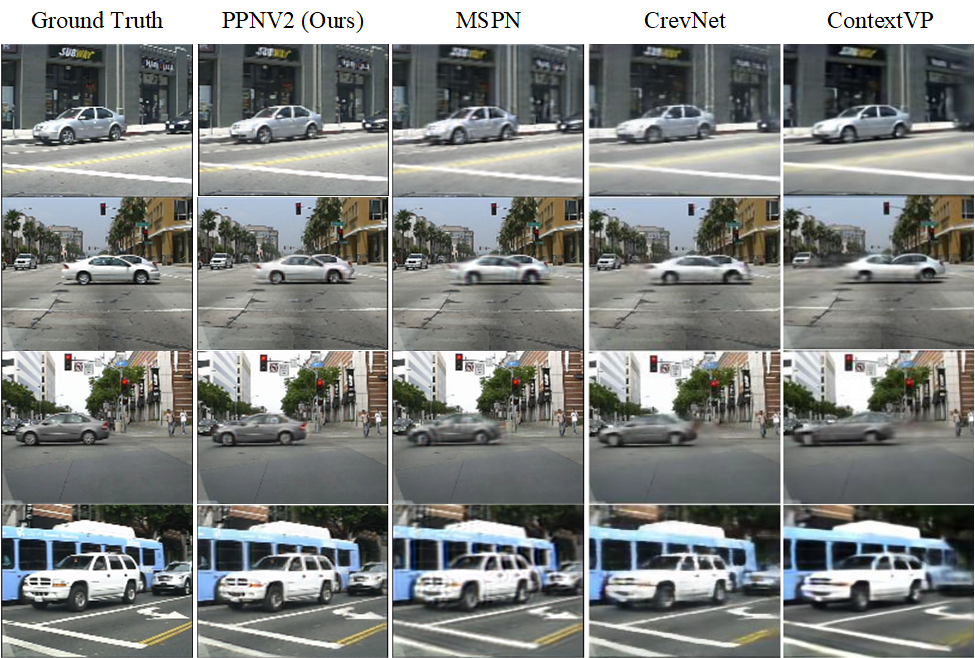}
	\end{minipage}
	\quad
	%\hfill
	\begin{minipage}{0.51\linewidth}
		\centering
		\vspace{-0.21cm}
		\renewcommand\arraystretch{1.47}
		\small
		\resizebox{\linewidth}{!}
		{
			\begin{tabular}{lcccccc}
				\hline
				\multirow{2}{*}{Methods} & \multicolumn{3}{c}{Caltech ($10 \rightarrow 15$)} & \multicolumn{3}{c}{KITTI ($10 \rightarrow 15$)} \\
				& SSIM    & PSNR    & LPIPS   & SSIM    & PSNR   & LPIPS  \\ \hline
				MCNet \cite{villegas2017decomposing}                   & 0.705   & -       & 37.34   & 0.555   & -      & 37.39  \\
				PredNet \cite{lotter2017deep}                 & 0.753   & -       & 36.03   & 0.475   & -      & 62.95  \\
				Voxel Flow   \cite{liu2017video}            & 0.711   & -       & 28.79   & 0.426   & -      & 41.59  \\
				Vid2vid \cite{wang2018video}                 & 0.751   & -       & 20.14   & -       & -      & -      \\
				FVSOMP \cite{wu2020future}                  & 0.756   & -       & 16.50   & 0.608   & -      & 30.49  \\
				PPNet \cite{ling2022pyramidal}                    & 0.812   & 21.3    & 14.83   & 0.617   & 18.24  & 31.07  \\
				MSPN \cite{ling2022predictive}                    & 0.818   & 23.88   & 10.98   & 0.629   & 19.44  & 32.10  \\
				CrevNet \cite{yu2020efficient}                    & 0.841   & -   & -  & -  & -  & - \\ \hline
				Ours             & \textbf{0.865}       & \textbf{25.44}       & \textbf{5.287}       & 0.621       & 19.32     & \textbf{15.45}     \\ \hline
			\end{tabular}
		}
	\end{minipage}
	\caption{The left shows the visualization examples of next predicted frames on Caltech dataset. The right shows the quantitative results on both Caltech and KITTI datasets, where the metrics are averaged over the 5 predicted frames. Zoom in for a better view.}
	\label{fig:Caltech}
	\vspace{-0.5cm}
\end{figure*}

Figure \ref{fig:KITTI} and \ref{fig:Caltech} present the results on Caltech and KITTI datasets. It is challenging to perform prediction on KITTI dataset because its scenes are much more complex and vary more drastically between frames, so most of the works \cite{lotter2017deep, ling2022pyramidal, ling2022predictive} usually predict blurry specious future frames, Nevertheless, it can be observed from Figure \ref{fig:KITTI} that our method can still generate believable results. The complexity of Caltech dataset is comparable to that of KITTI, but its inter-frame variation is much smaller.  According to the results, our method outperforms existing works both in pixel-wise accuracy and visualization performance, that is, it generates finer images and recovers more details.

\subsection{Ablations and comparisons}

\begin{figure*}[!t]
	\centering
	\begin{minipage}{0.50\linewidth}
		\centering
		\vspace{-0.30cm}
		\includegraphics[width=\linewidth]{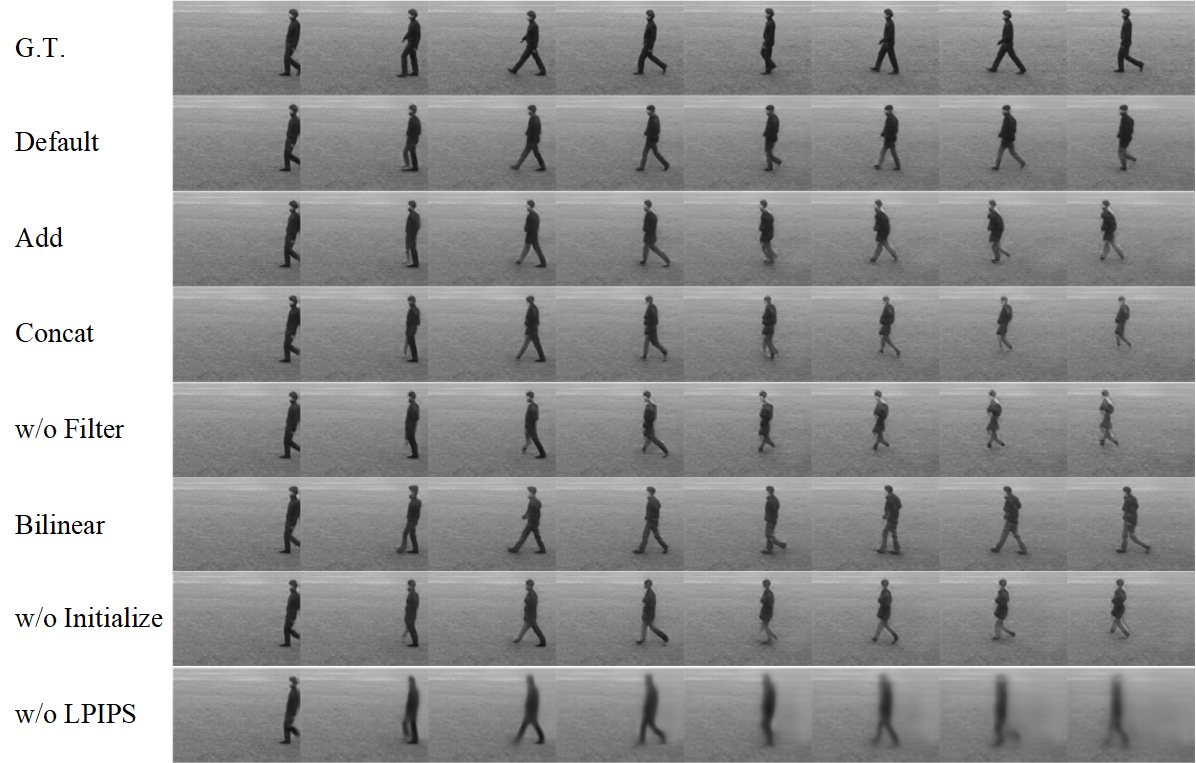}
	\end{minipage}
	\quad
	%\hfill
	\begin{minipage}{0.46\linewidth}
		\centering
		\vspace{-0.30cm}
		\renewcommand\arraystretch{2.38}
		\small
		\resizebox{\linewidth}{!}
		{
			\begin{tabular}{lcccccc}
				\hline
				\multirow{2}{*}{Ablation} & \multicolumn{3}{c}{$10 \rightarrow 20$} & \multicolumn{3}{c}{$10 \rightarrow 40$} \\
				& SSIM $\uparrow$  & PSNR $\uparrow$  & LPIPS $\downarrow$ 
				& SSIM $\uparrow$  & PSNR $\uparrow$  & LPIPS $\downarrow$   \\ \hline
				Default                   & \textbf{0.893}   & \textbf{32.05}  & \textbf{4.76}   & \textbf{0.833}   & \textbf{28.97}  & \textbf{8.93}   \\
				Add                       & 0.886   & 31.87  & 5.08   & 0.817   & 28.46  & 9.75   \\
				Concat                    & 0.888   & 31.92  & 5.16   & 0.820   & 28.71  & 9.56   \\
				w/o Filter                  & 0.882   & 31.70  & 5.21   & 0.808   & 28.39  & 9.71   \\
				Bilinear                  & 0.887   & 31.85  & 5.10   & 0.816   & 28.38  & 9.37   \\
				w/o Initialize                 & 0.883   & 31.62  & 5.22   & 0.813   & 28.41  & 9.61   \\
				w/o LPIPS                   & 0.889   & 31.99  & 11.45  & 0.824   & 28.79  & 20.41 \\ \hline
			\end{tabular}
		}
	\end{minipage}
	\caption{Ablations and comparisons on the KTH dataset. The left shows the visualization examples of 30 predicted frames. The right shows the quantitative results. Zoom in for a better view.}
	\label{fig:ablation}
	\vspace{-0.2cm}
\end{figure*}

We conducted several ablation studies to estimate the effects of each artifact or method proposed in this work. Each ablation study was performed on the basis of the default method with corresponding artifacts removed or replaced. Figure \ref{fig:ablation} presents the results obtained by training on KTH dataset. 

\begin{itemize}
\item First, we highlight the superiority of modulation model (Section 3.1) by comparing with the point-wise addition ``Add'' and dimension-wise concatenation ``Concat'' methods . To ensure the same number of parameters and computational overhead, we add an additional convolutional unit for the ``Add" approach. It can be observed that the proposed module outperforms the traditional approaches both in quantitative and qualitative evaluation. Furthermore, the ``Add'' and ``Concat'' approaches both experienced crashes (sudden drops in accuracy) during training, while the modulation module did not, indicating that the gradient propagation of this module is indeed more stable.

\item Second, whether the low-pass filter (Section 3.2) is used or not has a greater impact on the prediction effect. It can be observed that in the absence of a low-pass filter for anti-aliasing (``w/o Filter"), the model generate unnatural images, where the  portrayal of the character is incoherent and the details are missing.

\item Third, we compare the interleaved upsampling method proposed in Section 3.3 with traditional bilinear upsampling (``Bilinear"). It seems that using only bilinear upsampling can also get good performance. But on closer inspection, our method is more coherent in depicting the color of the character while the bilinear upsampling produces choppy grayish white spots.

\item Fourth, we further perform ablation (``w/o Initialize") on the proposed initialization described in Section 3.3. It is obvious that using the proposed initialization achieves better results than common methods such as ``Kaiming uniform" \cite{he2015delving} with the same number of training epoch.

\item Fifth, we remove the LPIPS loss and only use Euclidean distance to train the model (``w/o LPIPS"). According to the results,  although its pixel-wise accuracy (SSIM and PSNR) is higher than the other ablation studies, its visual performance is worse, which further proves our previous conjecture: using only deterministic loss may produce specious results. It is necessary to use deep features to calculate loss to obtain results that are more in line with the human visual system.
\end{itemize}

\section{Limitations and future work}
In this work, we propose to define learnable cutoff frequency for low-pass filter to suppress aliasing in neural networks. Unfortunately, the detailed reasons why the network chooses such a ratio coefficient have not been explored well. We suspect that this is mainly depended on the size of the input images. For example, the Caltech (size of $128 \times 160$) dataset has similar scenarios to KITTI (size of $128 \times 256$), but the ratio coefficients learned are quite different (Table 1, Level \uppercase\expandafter{\romannumeral4}). This may limit the model to achieve good performance on tasks with different input sizes. Moreover, anti-aliasing is not suitable for some special scenarios such as datasets with only black and white pixels. Low-pass filtering can severely blur the boundary and produce  unwanted pixels. Therefore, it is better to remove the low-pass filter while performing on this kind of datasets. 

In the future, we will employ filters more carefully and explore additional applicable scenarios. For instance, a learnable band-pass filter may be utilized to extract signals in specific frequency bands of interest, mitigating the interference from extraneous information during calculation. Additionally, using LPIPS to calculate the loss for training plays an important role in this work, however, its representation ability could still be enhanced since the authors only employ simple backbones such as VGG and AlexNet. It might be interesting to train a more powerful neural network which would enable the generator network to generate accurate and clear future frames without using any deterministic losses. Finally, the present network architecture lacks control over the update frequency of neurons at different levels. We will further improve the architecture, with the goal of making it possible to adjust the update frequency of higher-level neurons according various conditions such as the scene to be predicted, the variations between frames and the prediction error.

\bibliographystyle{plain}
%%%%%%%%%%%%%%%%%%%%%%%%%%%%%%%%%%%%%%%%%%%%%%%%%%%%%%%%%%%%
\bibliography{references}

\clearpage

\appendix
\section{Additional results}
\begin{figure*}[!h]
	\centering
	\begin{minipage}{0.45\linewidth}
		\centering
		%\vspace{-0.30cm}
		%\setlength{\abovecaptionskip}{0.28cm}
		\includegraphics[width=\linewidth]{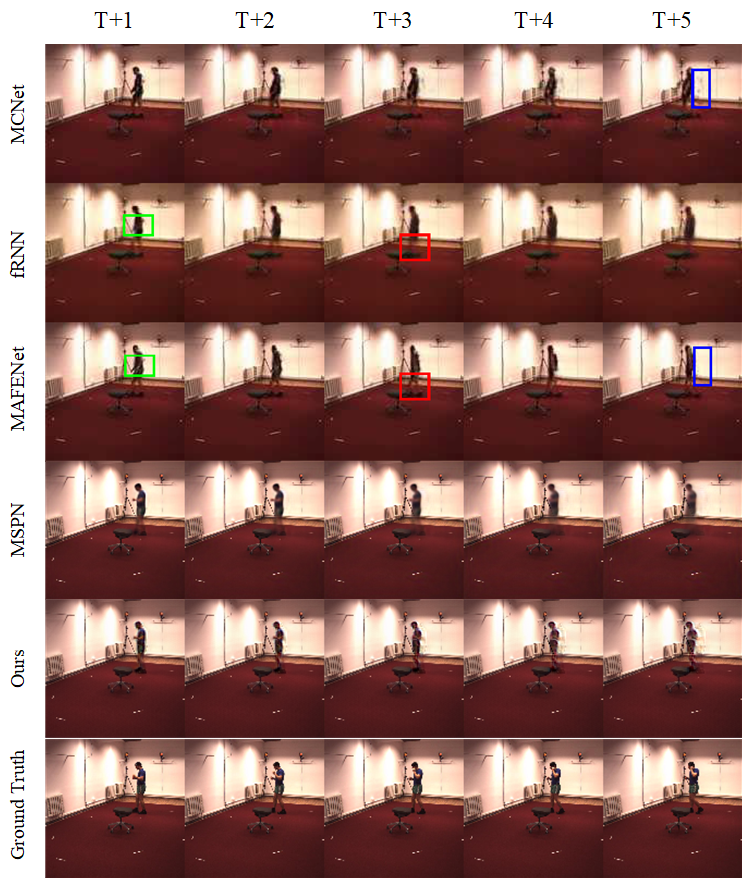}
	\end{minipage}
	\quad
	%\hfill
	\begin{minipage}{0.51\linewidth}
		\centering
		\vspace{0.31cm}
		\renewcommand\arraystretch{1.77}
		\small
		\resizebox{\linewidth}{!}
		{
			\begin{tabular}{lcccccc}
				\hline
				Methods                      & Metric & T=2    & T=4    & T=6    & T=8    & T=10   \\ \hline
				\multirow{3}{*}{MCNet \cite{villegas2017decomposing} }       & PSNR   & 30.0   & 26.55  & 24.94  & 23.90  & 22.83  \\ 
				& SSIM   & 0.9569 & 0.9355 & 0.9197 & 0.9030 & 0.8731 \\
				& LPIPS  & 0.0177 & 0.0284 & 0.0367 & 0.0462 & 0.0717 \\
				\hline
				\multirow{3}{*}{fRNN \cite{oliu2018folded}}        & PSNR   & 27.58  & 26.10  & 25.06  & 24.26  & 23.66  \\
				& SSIM   & 0.9000 & 0.8885 & 0.8799 & 0.8729 & 0.8675 \\
				& LPIPS  & 0.0515 & 0.0530 & 0.0540 & 0.0539 & 0.0542 \\
				\hline
				\multirow{3}{*}{MAFENet \cite{lin2020motion}}     & PSNR   & 31.36  & 28.38  & 26.61  & 25.47  & 24.61  \\
				& SSIM   & 0.9663 & 0.9528 & 0.9414 & 0.9326 & 0.9235 \\
				& LPIPS  & 0.0151 & \textbf{0.0219} & \textbf{0.0287} & 0.0339 & 0.0419 \\
				\hline
				\multirow{3}{*}{MSPN \cite{ling2022predictive}} & PSNR   & 31.95  & 29.19  & 27.46  & 26.44 & 25.52   \\
				& SSIM & \textbf{0.9687} & \textbf{0.9577} & 0.9478 & 0.9382 & 0.9293 \\
				& LPIPS  & \textbf{0.0146} & 0.0271 & 0.0384 & 0.0480 & 0.0571 \\
				\hline
				\multirow{3}{*}{Ours} & PSNR   & \textbf{32.07}  & \textbf{30.08}  & \textbf{28.81}  & \textbf{28.12}  & \textbf{27.55}   \\
				& SSIM & 0.9645 & 0.9566 & \textbf{0.9510} & \textbf{0.9461} & \textbf{0.9421} \\
				& LPIPS  & 0.0169 & 0.0239 & 0.0288 & \textbf{0.0337} & \textbf{0.0381} \\
				\hline
			\end{tabular}
		}
	\end{minipage}
	\caption{Qualitative and quantitative results on the Human3.6M. Zoom in for a better view.}
	\label{fig:Human3.6M}
	
\end{figure*}

Human3.6M and Moving MNIST are another popular datasets for video prediction. The results of our proposed method on the Human3.6M dataset are presented in Figure \ref{fig:Human3.6M}. It can be observed from the quantitative results that our proposed method outperform other works in terms of longer-term prediction. According to the visualization examples, the characteristic of this dataset is that the background is invariant, and all we need to predict is the actions of the human in the video sequence. It seems oversimplified, and many works have also achieved pretty good quantitative results. However, most of the works usually obtain pixel-wise accurate but specious results by only replicating the background. In contrast, while these works gradually blur the human in images, our model can still better recover the human's silhouette and predict the motion. This may explain why our method can achieve better longer-term prediction performance.

\begin{figure*}[!h]
	\vspace{0.5cm}
	\centering{\includegraphics[width=1.0\textwidth]{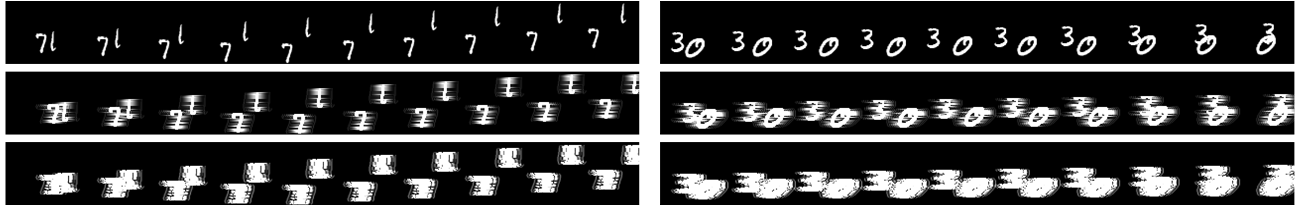}}
	\caption{The results obtained after low-pass filtering the Moving MNIST dataset.The first row indicates the original data. The second row is the results obtained by setting the ratio coefficient $2f_c/f_s$ of the low-pass filter to 0.99, while the third row is obtained by setting to 0.8.}
	\label{fig:Filtered_MNIST}
\end{figure*}

\begin{figure*}[!h]
	\vspace{0.50cm}
	\centering
	\begin{minipage}{0.45\linewidth}
		\centering

		\includegraphics[width=\linewidth]{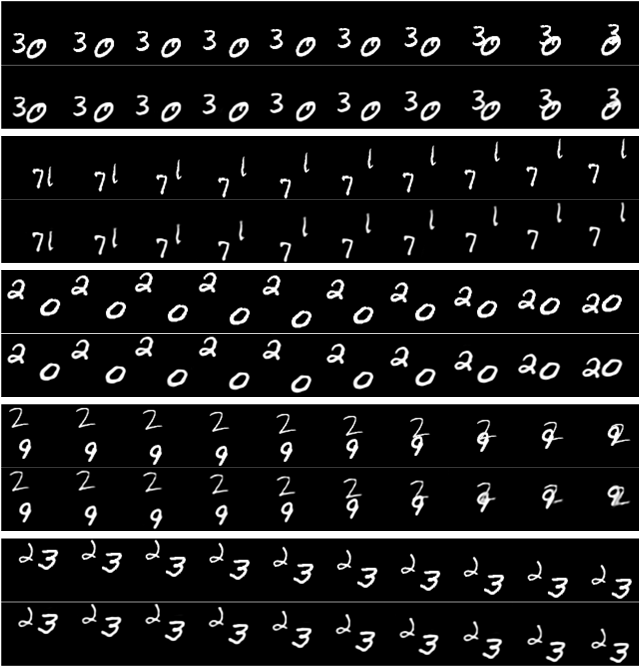}
	\end{minipage}
	\quad
	%\hfill
	\begin{minipage}{0.51\linewidth}
		\centering
		%	\vspace{0.31cm}
		\renewcommand\arraystretch{1.245}
		\small
		\resizebox{\linewidth}{!}
		{
			\begin{tabular}{lcc}
				\hline
				\multirow{2}{*}{Methods} & \multicolumn{2}{c}{10 $\rightarrow$ 20}                               \\
				& \multicolumn{1}{c}{SSIM $\uparrow$} & \multicolumn{1}{c}{MSE $\downarrow$} \\ 
				\hline
				2D ConvLSTM \cite{shi2015convolutional}             & 0.763                    & 82.2                    \\
				PredRNN++ \cite{wang2018predrnn++}                 & 0.870                    & 47.9                    \\
				E3D-LSTM \cite{wang2018eidetic}                 & 0.910                    & 41.3                    \\
				Variational 2D ConvLSTM \cite{chung2015recurrent}  & 0.816                    & 60.7                    \\
				Improved VRNN \cite{castrejon2019improved}           & 0.776                    & 129.2                   \\
				Variational 3D ConvLSTM \cite{razali2021log} & 0.896                    & 39.4                    \\
				Conv-TT-LSTM  \cite{su2020convolutional}            & 0.915                    & 53.0                    \\
				LMC-Memory  \cite{lee2021video}               & 0.924                    & 41.5                    \\
				MAU \cite{chang2021mau}            & 0.937        & 27.6          \\
				SimVP \cite{gao2022simvp}            & 0.948        & 23.8          \\
				CrevNet \cite{yu2020efficient}            & 0.949        & \textcolor{blue}{22.3}          \\
				TAU \cite{tan2022temporal}            & \textcolor{red}{0.957}        & \textcolor{red}{19.8}         \\
				\hline
				Ours             & \textcolor{blue}{0.950}      & 23.4   \\ \hline                
			\end{tabular}
		}
	\end{minipage}
	\caption{Results on the Moving MNIST dataset. Left: the visualization examples, in each group, the first row indicates the ground truth, the second row indicates the prediction. Right: the quantitative results.}
	\label{fig:MNIST}
	
\end{figure*}

\begin{figure*}[!h]
	\centering{\includegraphics[width=1.0\textwidth]{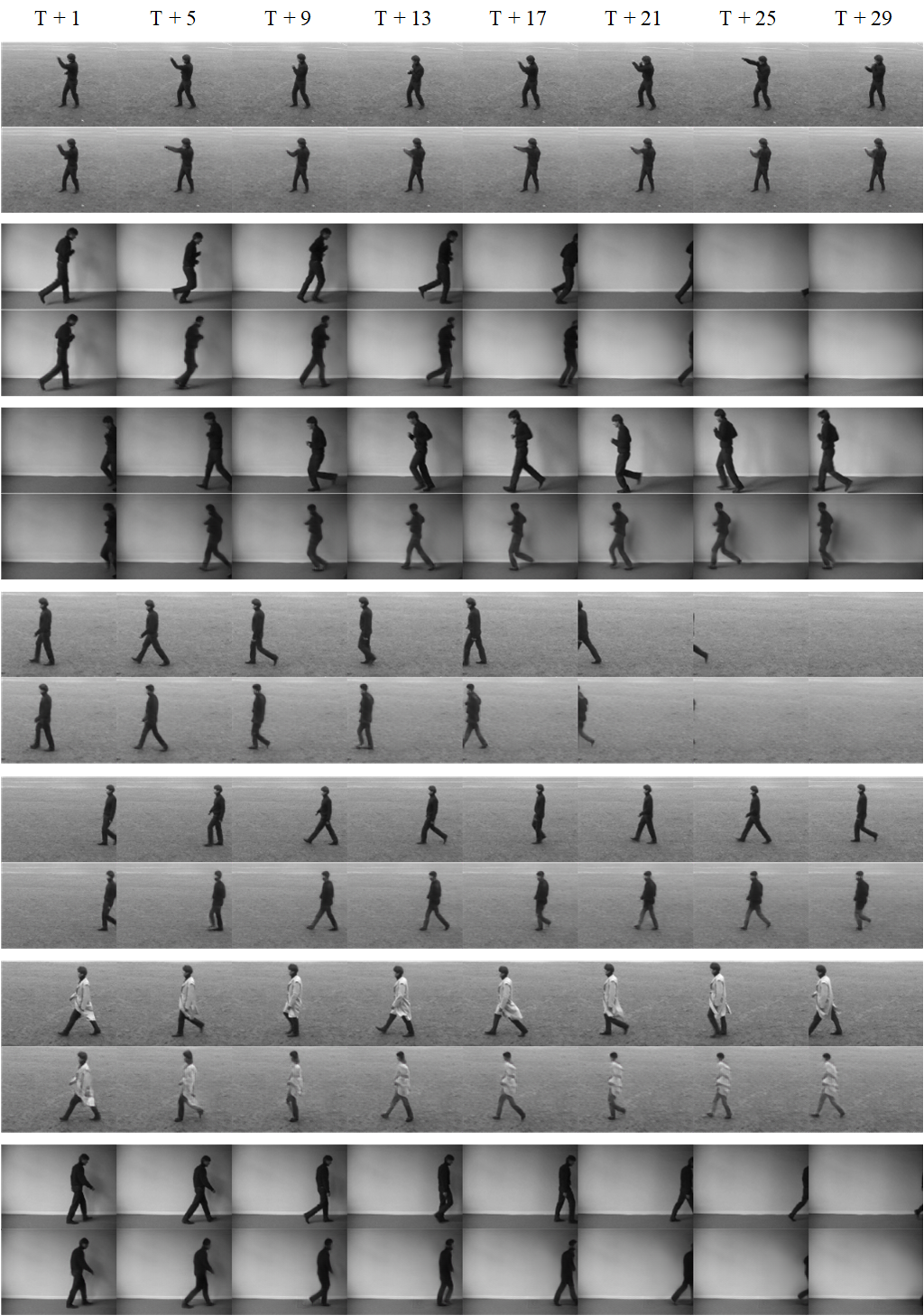}}
	%\vspace{-0.5cm}
	\caption{Additional visualization results on the KTH dataset. We use 10 frames to predict next 30 frames. In each group, the first row indicates the ground truth, the second row indicates the prediction.}
	\label{fig:additional_KTH}
\end{figure*}

\begin{figure*}[!h]
	\centering{\includegraphics[width=1.0\textwidth]{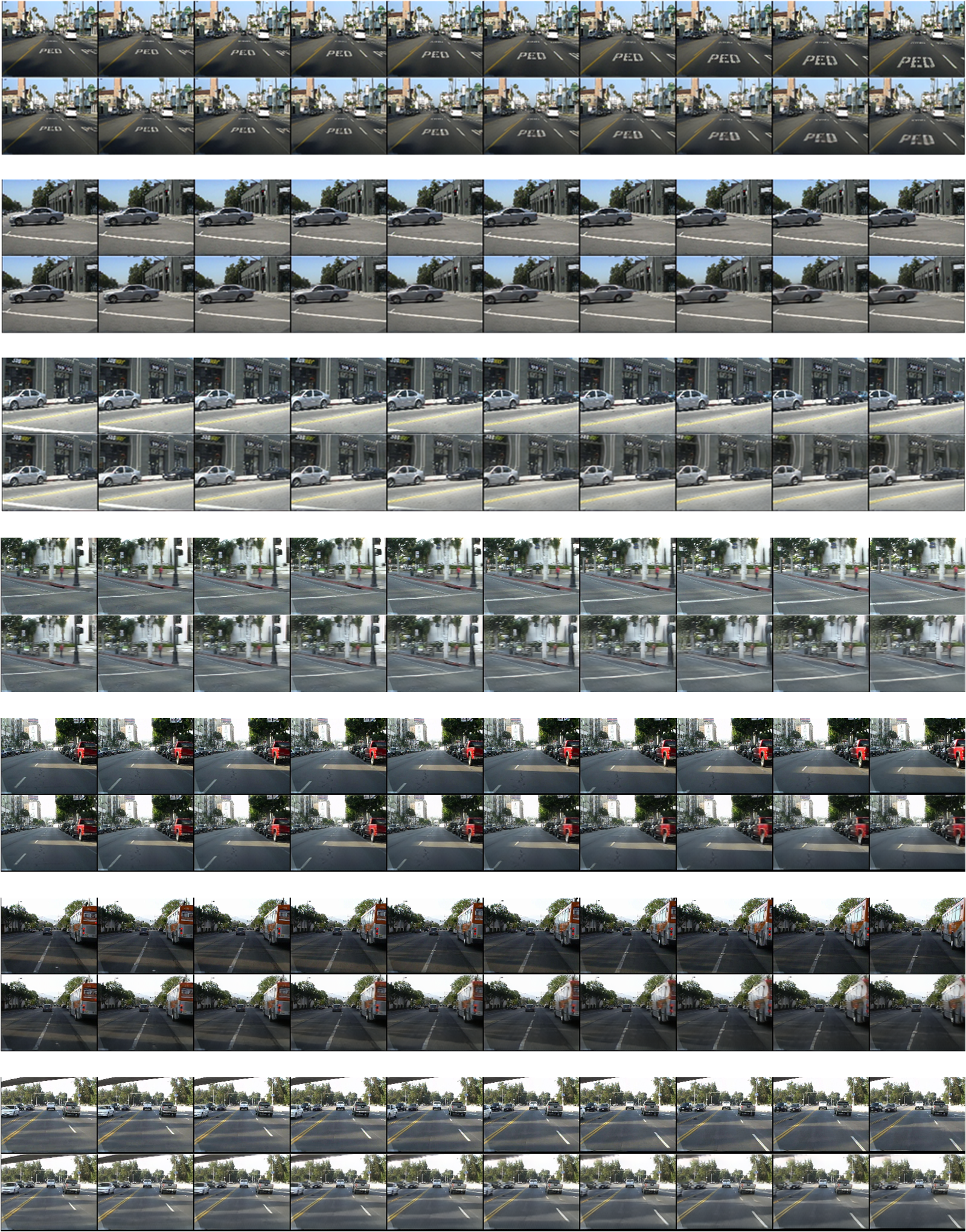}}
	%\vspace{-0.5cm}
	\caption{Additional visualization examples on the Caltech dataset. We use 10 frames to predict next 10 frames. In each group, the first row indicates the ground truth, the second row indicates the prediction.}
	\label{fig:additional_caltech}
\end{figure*}

The Moving MNIST dataset is a popular synthetic dataset for video prediction tasks, known for its simple scenes and events. However, our methods focus more on predicting natural scenes, which requires several modifications to achieve comparable performance on this dataset. Firstly, we must remove the low-pass filters in our network. As explained in Section 6 of the main submission, anti-aliasing is not suitable for datasets with only black and white pixels, as low-pass filtering can severely blur the boundary. As shown in Figure \ref{fig:Filtered_MNIST}, even if we select a ratio coefficient as high as 0.99, it will also greatly blur the boundary. Secondly, we have to forgo calculating the LPIPS loss, as the pre-trained model has not been trained on this kind of dataset and, therefore, cannot guide the learning of the generator network. The final results are presented in Figure \ref{fig:MNIST}, we can also obtain predictive performance comparable to existing works. Additional visualization examples on the KTH and Caltech datasets are shown in Figure \ref{fig:additional_KTH} and \ref{fig:additional_caltech}, respectively.

\clearpage

\section{Detailed calculation of the model}

We would like to describe the detailed calculation of our model based on Figure \ref{fig:TotalNet}, which shows the overall architecture, detailed network modules and calculations of our model. Firstly, we will introduce the general computations of neurons at level $l$ and time-step $t$, where $0 \le l \le L-1$ and $0 \le t \le T-1$, $L$ denotes the number of network level and $T$ denotes the total length of input sequence. In the following descriptions, we will use $f_{name}^l(\cdot)$ to represent the calculation of network module named $name$ at level $l$.

The far right of Figure \ref{fig:TotalNet} shows a more detailed calculation process. In general, the neurons need to combine the following input signals to calculate the final prediction: $f_l^t$, $E_l^t$, $E_{l-1}^t$ and $P_{l+1}^t$. The $f_l^t$ and $E_l^t$ represent local sensory input and prediction error, where $f_l^t$ is computed by using the proposed downsampling artifact ($DF_f$) to perform feature extraction and downsampling on the lower-level sensory input $f_{l-1}^t$ (if $l = 0$, then $f_l^t = x_t$, where $x_t$ denotes the video frame at time-step $t$). The sensory input $f_l^t$ is an important signal, as the neurons at this level are unable to perform any computations in its absence.
\begin{equation}
	f_l^t = f_{DS_f}^{l-1}(f_{l-1}^t)
\end{equation}
And, the $E_l^t$ is obtained by performing point-wise subtraction between current sensory input $f_l^t$ and previous prediction $P_l^{t-1}$. Similar to PPNet \cite{ling2022pyramidal} and PredNet \cite{lotter2017deep}, we calculate the positive and negative errors respectively and concatenate them together by channels
\begin{equation}
	E_l^t = [|f_l^t - P_l^{t-1}|; |P_l^{t-1} - f_l^t|]
\end{equation}

\begin{figure*}[!b]
	
	\centering{\includegraphics[width=1.0\textwidth]{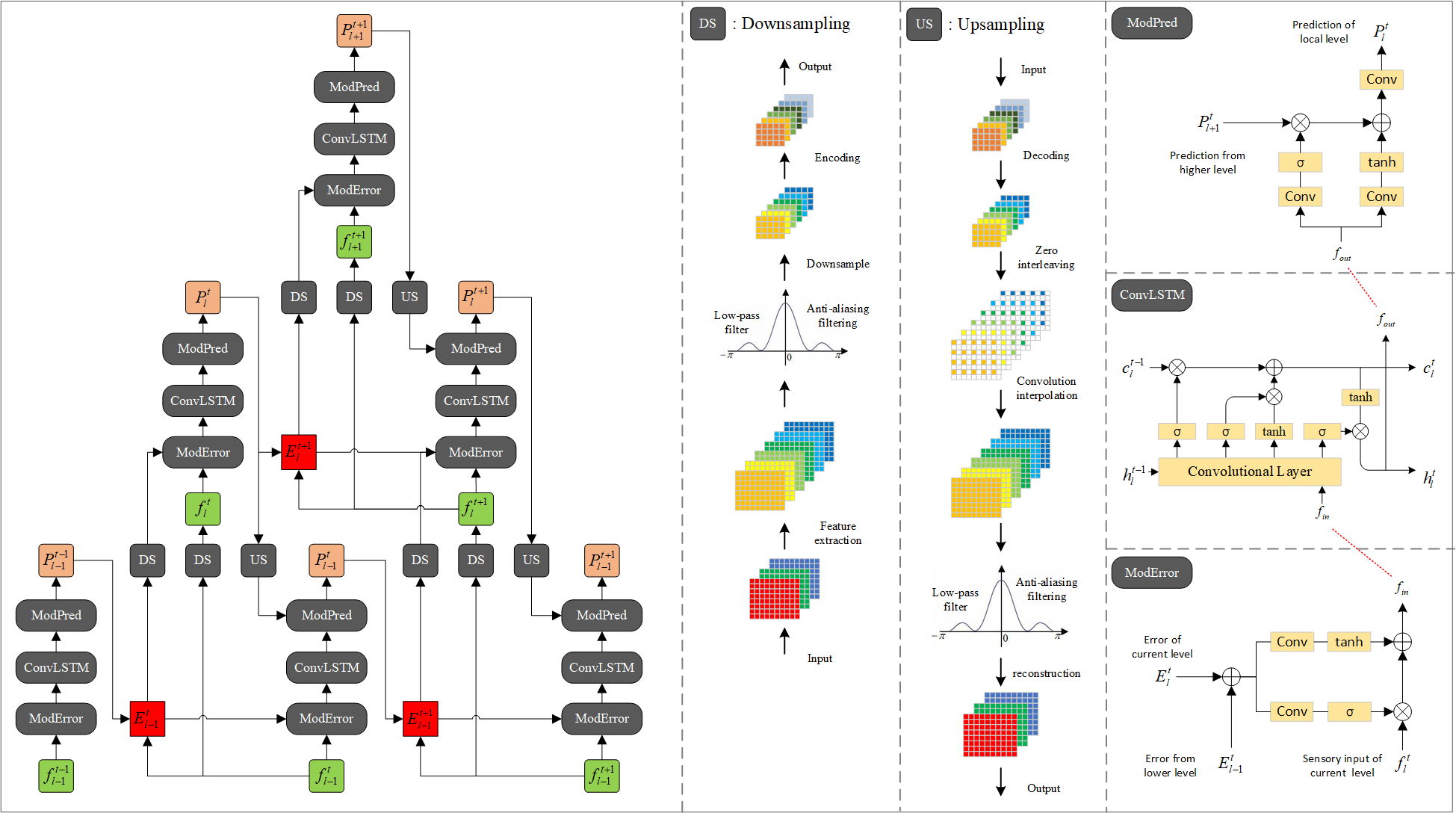}}
	%\vskip -0.05in
	\caption{Overview of our model. From left to right are the overall framework, the downsampling artifact, the upsampling artifact and the detailed calculation process. "ModError" and "ModPred" are special modulation modules. $f_l^t$, $P_l^t$ and $E_l^t$ represent the sensory input, prediction and prediction error at time-step $t$ and level $l$, respectively. Zoom in for a better view.}
	
	\label{fig:TotalNet}
	\vskip -0.1in
\end{figure*}
The $E_{l-1}^t$ denotes the lower-level prediction error. According to the predictive coding framework, it will be propagated upward to higher level by performing calculation ($f_{DS_{E}}^{l-1} (E_{l-1}^t)$) with another downsampling artifact ($DS_E$). Then, we perform point-wise addition with local prediction error $E_l^t$ to update the representations
\begin{equation}
	E_l^t \leftarrow \alpha E_l^t + \beta f_{DS_{E}}^{l-1} (E_{l-1}^t)
\end{equation}
where the reason we choose to perform point-wise addition is that they both represent the prediction error (if $l=0$, then $E_l^t \leftarrow E_l^t$ since there is no prediction error from lower-level network).  $\alpha$ and $\beta$ are learnable coefficients that enable the model to assign appropriate weights through training. Next, we fuse the prediction error $E_l^t$ and sensory input $f_l^t$ with the proposed modulation module $ModError$, to calculate the input $f_{in}$ for $ConvLSTM$ module
\begin{equation}
	f_{in} = f_{ModError}^l (f_l^t, E_l^t)
\end{equation}
Then, the input $f_{in}$ is combined with the memory states $c_l^{t-1}$ and $h_l^{t-1}$ to calculate the output $f_{out}$, which is considered preliminary prediction
\begin{equation}
	f_{out}, c_l^t, h_l^t = f_{ConvLSTM}^l (f_{in}, c_l^{t-1}, h_l^{t-1})
\end{equation}
Finally, we combine the output $f_{out}$ and prediction $P_{l+1}^t$ from higher level to calculate the local prediction $P_l^t$, where the higher-level prediction $P_l^t$  are first upsampled and reconstructed to the lower-level features ($f_{DS}^{l+1} (P_{l+1}^t)$) using the proposed upsampling artifact $DS$. Then we perform another modulation calculation $ModPred$ to obtain the final prediction $P_l^t$ (if $l = L-1$, then $P_l^t = f_{out}$, that is, at the highest level, there is no prediction from higher level)
\begin{equation}
	P_l^t = f_{ModPred}^l (f_{US}^{l+1} (P_{l+1}^t), f_{out})
\end{equation}

The general calculations of neurons at each level are described above. For more details, please refer to our code: \url{https://github.com/ANNMS-Projects/PNVP}

\section{Training details}
We ran all experiments using 4 TITAN Xp and 4 RTX3090 GPUS, PyTorch 1.12.1, CUDA 11.4. The entire project lasted about eight months, and it took us about five days to run an experiment with the KTH dataset on a single TITAN Xp GPU. Many of our hyperparameters, including learning rate, weighting factors of losses and setting of optimizer are described below:

\begin{itemize}
	\item Optimizer: we use Adam optimizer with $\beta_1 = 0$, $\beta_2 = 0.99$, and $\epsilon = 10^{-8}$
	
	\item Learning rate: its value is set between $10^{-4}$ and $10^{-3}$, which may be adjusted depending on the training dataset.
	
	\item $\lambda_t$: the weighting coefficient by time, described as Eq.7 in the main paper. Influenced by the initialization, we downweight the loss for the first timestep ($\lambda_0 = 0.5$), while the others are set to the same value: $\lambda_t = 1, 1 \le t \le T-1$. We have tried to force the network to reduce the error in longer-term prediction as much as possible by gradually increasing the weighting coefficient over time, to obtain better long-term prediction performance. However, it turns out that early predictions are equally important, in which even small errors can make subsequent predictions worse since the model is calculated in a Markov chain fashion.
	
	\item $\lambda_l$: the weighting coefficient by level, described as Eq.7 in the main paper. It decreases linearly with the network level increases: $\lambda_l = \frac{(L-1) - l}{L - 1}, 0 \le l \le L-1$. What we ultimately require is pixel-level predictions, so we don't try to force the network to predict the same representations as ground truth at higher-levels. Larger gaps are allowed at higher level networks, but we hope that these gaps will be reduced in the process of gradually restoring lower-level representations.
	
	\item $\lambda$: the weighting coefficient for LPIPS loss, described as Eq.8 in the main paper. It is used to increase the proportion of LPIPS loss and avoid the gradients to vanish, which is calculated by: $\lambda = c \times h \times w$, to ensure the same size as the other two Euclidean distance losses. The $c, h, w$ represent the channels, height and width of the input images, respectively.
\end{itemize}

\end{document}